\documentclass[lettersize,journal]{IEEEtran}
\usepackage{amsmath,amssymb,amsfonts}
\usepackage{algorithmic}
\usepackage{algorithm}
\usepackage{array}
\usepackage{textcomp}
\usepackage{stfloats}
\usepackage{url}
\usepackage{verbatim}
\usepackage{hyperref}
\usepackage{graphicx}
\usepackage{color}
\usepackage{subfigure}
\usepackage{amsthm}
\usepackage{thmtools, thm-restate}
\usepackage{cite}
\usepackage{multirow}
\begin{document}

\title{Affinity Uncertainty-based Hard Negative Mining in Graph Contrastive Learning}
\author{{Chaoxi Niu, Guansong Pang,~\IEEEmembership{Member,~IEEE}, Ling Chen,~\IEEEmembership{Senior Member,~IEEE}}
\thanks{Chaoxi Niu and Ling Chen are with the Australian Artificial Intelligence Institute, University of Technology Sydney, Sydney, NSW 2007, Australia. (email:~Chaoxi.Niu@student.uts.edu.au; Ling.Chen@uts.edu.au).

Guansong Pang is with School of Computing and Information Systems, Singapore Management University, 178902, Singapore. 

Corresponding author: G. Pang (email:~pangguansong@gmail.com).
}}

\markboth{Journal of \LaTeX\ Class Files,~Vol.~14, No.~8, August~2021}%
{Shell \MakeLowercase{\textit{et al.}}: A Sample Article Using IEEEtran.cls for IEEE Journals}

\IEEEpubid{0000--0000/00\$00.00~\copyright~2021 IEEE}

\maketitle

\begin{abstract}
Hard negative mining has shown effective in enhancing self-supervised contrastive learning (CL) on diverse data types, including graph contrastive learning (GCL). Existing hardness-aware CL methods typically treat negative instances that are most similar to the anchor instance as hard negatives, which helps improve the CL performance, especially on image data. However, this approach often fails to identify the hard negatives but leads to many false negatives on graph data. This is mainly due to that the learned graph representations are not sufficiently discriminative due to over-smooth representations and/or non-i.i.d. issues in graph data. To tackle this problem, this paper proposes a novel approach that builds a discriminative model on \textit{collective affinity} information (i.e., two sets of pairwise affinities between the negative instances and the anchor instance) to mine hard negatives in GCL. In particular, the proposed approach evaluates how confident/uncertain the discriminative model is about the affinity of each negative instance to an anchor instance to determine its hardness weight relative to the anchor instance. This uncertainty information is then incorporated into existing GCL loss functions via a weighting term to enhance their performance. The enhanced GCL is theoretically grounded that the resulting GCL loss is equivalent to a triplet loss with an \textit{adaptive} margin being exponentially proportional to the learned uncertainty of each negative instance. Extensive experiments on 10 graph datasets show that our approach i) consistently enhances different state-of-the-art GCL methods in both graph and node classification tasks, and ii) significantly improves their robustness against adversarial attacks. Code is available at \renewcommand\UrlFont{\color{blue}\tt}
\url{https://github.com/mala-lab/AUGCL}.
\end{abstract}


\section{Introduction}
\IEEEPARstart{G}{raph} is ubiquitous and plays an important role in various fields \cite{xia2021survey, ren2023graph, wu2020comprehensive}, such as social networks, bioinformatics, chemistry, etc. Due to its non-Euclidean nature, learning expressive graph representations is one crucial foundation of different graph mining tasks, such as graph classification and node classification. In recent years, graph neural networks (GNNs) have become predominant in achieving this goal. Most existing GNNs focus on supervised or semi-supervised learning settings \cite{xu2018how, kipf2016semi, velivckovic2018graph}, where class label information is required for training the GNNs. However, obtaining such information is hard or costly, especially for graph data which is at large scale and/or demands strong domain knowledge to accurately perform the data annotation. Recently, self-supervised learning of GNNs \cite{xie2022self,liu2021graph} which can learn graph representations without accessing ground truth labels was introduced to tackle this issue and has attracted significant research interests.
\IEEEpubidadjcol

Graph contrastive learning (GCL) has become one of the most popular self-supervised methods for graph representation learning \cite{velickovic2019deep, hassani2020contrastive, sun2019infograph, you2020graph, zhu2021graph, you2021graph, xia2022simgrace, qiu2020gcc, wan2023self, chen2022subgraph, zheng2022gzoom}. It focuses on learning representations by maximizing the mutual information between augmentations of the same instance, in which the augmentations of the same graph/node are often treated as positive instances, with the other graphs/nodes as negative instances \cite{xie2022self,liu2021graph}. 

Despite the impressive successes achieved by current GCL methods, their learning capability can be largely limited by the way they choose negative samples \cite{kalantidis2020hard,robinson2020hard,chuang2020debiased, xia2022progcl}. One commonly used negative selection approach is to randomly select negative instances from a sufficiently large batch or a memory bank, and then treat all negative instances equally in contrastive learning. However, this approach typically overlooks the negative instances that can provide more information for contrastive learning than the others. These informative negative instances are commonly referred to as \textit{hard negative} instances. Many prior studies \cite{kalantidis2020hard, robinson2020hard, xia2022progcl} have demonstrated the critical importance of hard negative instances over counterparts (e.g., easy negatives that are distant from the positive in both semantics and representations) in learning discriminative features and enabling fast convergence.

Many recent contrastive learning (CL) methods \cite{chuang2020debiased, robinson2020hard, kalantidis2020hard, wu2020conditional, lee2021imix} thus incorporate hard negative mining methods into their training process to leverage these hard negative instances. These methods typically treat negative instances that are most similar to the anchor instance as the hard negatives \cite{chuang2020debiased, robinson2020hard, kalantidis2020hard, wu2020conditional, lee2021imix}. Although achieving improved CL performance on image data, these hard negative mining approaches often perform poorly on graph data, as shown in some recent studies \cite{zhu2021empirical, xia2022progcl} and our experiments. This is mainly because the learned graph representations are not sufficiently discriminative due to i) the non-i.i.d. (non-independent and identically distributed) nature of graph data, e.g., nodes with the same label tend to be densely connected in graph data, and ii) the over-smooth graph representations resulting from the iterative message passing mechanism. Consequently, for graph data, the most similar negatives to the anchor can be false negatives with high probability. To address this issue, the very recent method ProGCL \cite{xia2022progcl} imposed a beta mixture model on the pairwise similarities between the negatives and the anchor to estimate the probability of a negative being true one, and it subsequently combined the estimated probability and the pairwise similarity to measure the hardness of the negatives. The method relies on the prior that the similarity distribution of negatives w.rt. positive is bimodal and works well in node classification tasks. It fails to work when its prior is not fully met. As shown in our experiments (Table~\ref{results} in Sec. \ref{sec:graph_node_results}), such failure cases occur in most graph classification datasets where ProGCL has very marginal improvement, or even worse performance, compared to the original GCL methods. Besides, AdaS \cite{adas2023} emphasized the contrastive learning on the negatives that are neither hard nor easy to discriminate from the anchor. HSAN \cite{HSAN} exploited hard samples by utilizing high-confidence clustering and comprehensive similarity information. However, HSAN is primarily proposed for node clustering. This work aims to eliminate the need to specify the similarity distribution prior as in ProGCL for better applicability in practice.
 
\begin{figure}
    \centering
    \includegraphics[width=0.48\textwidth]{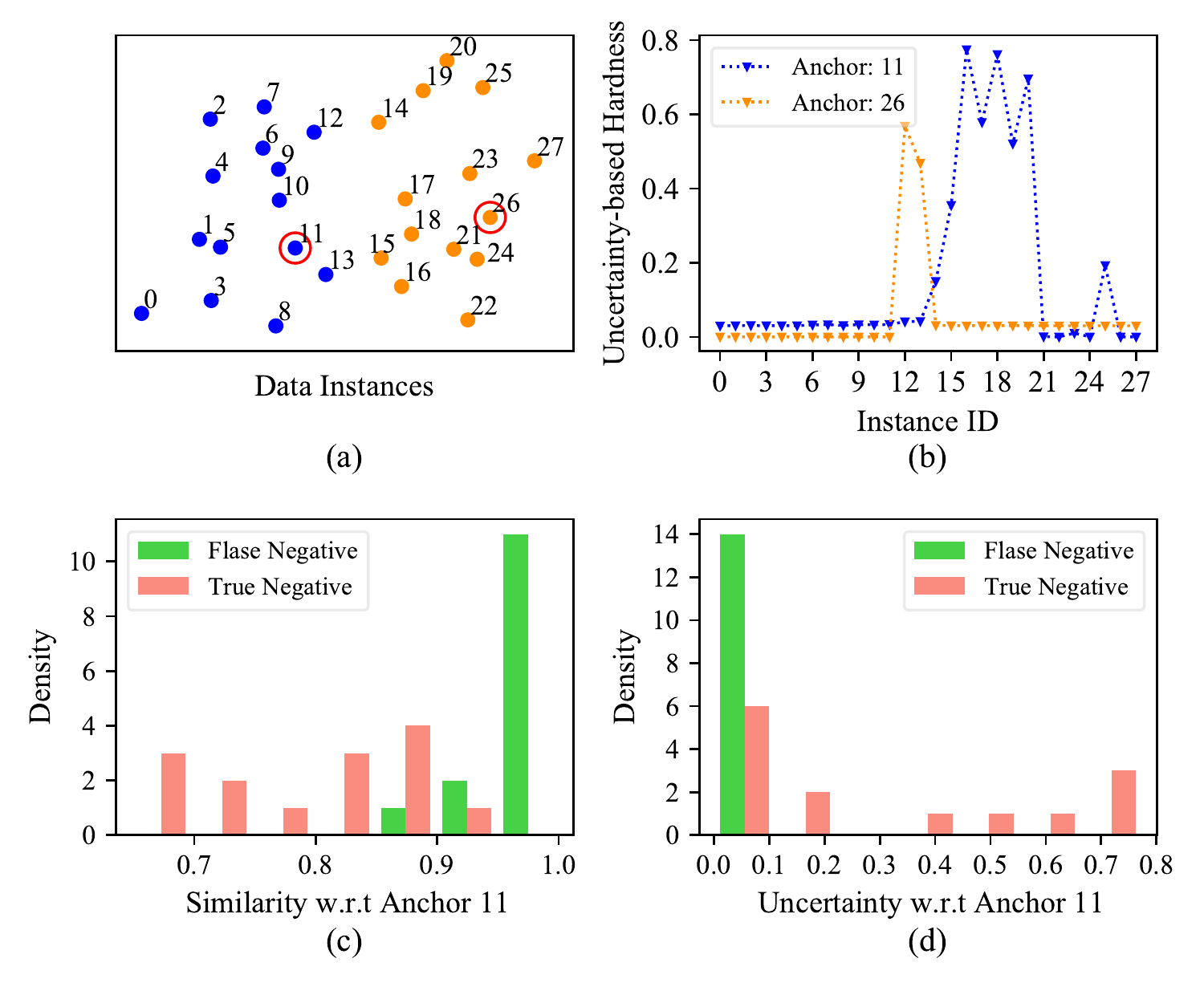}
    \caption{(\emph{a}): Two groups of data instances in blue and orange. (\emph{b}): The affinity uncertainty-based hardness results learned by our approach using instance 11 or 26 as the anchor instance. Instances with a larger uncertainty are more likely to be hard negative samples w.r.t. the anchor instance. (\emph{c}): The histograms of the similarity of the instances to the anchor instance 11. It is clear that treating the most similar instances to the anchor as the hard negatives can lead to many false negatives. (\emph{d}): The uncertainty results learned by our approach for the instances w.r.t the anchor instance 11, where true negatives including hard negatives have large uncertainty values (and thus large hardness weights) while false negative cases receive very small uncertainty values.}
    \label{fig:uncervisual1}
\end{figure}

To this end, this paper introduces a novel \underline{A}ffinity \underline{U}ncertainty-based hard negative mining for \underline{G}raph \underline{C}ontrastive \underline{L}earning, termed as AUGCL, to tackle this problem. AUGCL learns a data-driven, affinity-based uncertainty estimator to evaluate the hardness of negative instances relative to each anchor instance, meaning that the hardness of an instance is dependent on the given anchor instance, as shown by an example in Fig. \ref{fig:uncervisual1}(a-b). Particularly, AUGCL builds a discriminative model on \textit{collective affinity} information (i.e, two sets of pairwise affinities between the negative instances and the anchor instance) to evaluate how confident/uncertain the discriminative model is about the affinity of each negative instance to the anchor instance. Instances that have a larger affinity uncertainty would be more likely to be hard negatives, and they are subsequently assigned with a larger hard-negative weight to receive more attention from the GCL models. By doing so, AUGCL learns discriminative affinity uncertainties for the negative instances relative to each anchor instance, as shown by the results of the anchor instance 11 in Fig. \ref{fig:uncervisual1}(b) and (d), where small and large uncertainty-based hardness values are assigned to false negatives and true negatives, respectively. By contrast, the current similarity-based methods that regard the most similar negative instances to the anchor instance as hard negatives fail to identify the truly hard negatives but lead to many false negatives, as shown in Fig. \ref{fig:uncervisual1}(c). Those learned hardness results can then be seamlessly incorporated into popular GCL models (e.g., InfoNCE-based models \cite{van2018representation}) as a hardness weight to enhance their performance. AUGCL addresses a similar issue as ProGCL, but it eliminates the prior information posited in ProGCL, enabling AUGCL to work more effectively on diverse node-level and graph-level datasets.

In summary, this work makes the following three main contributions. 
\begin{itemize}
    \item We propose a novel approach AUGCL that utilizes the modeling of collective affinities to take account of the non-i.i.d. and over-smooth representations issues in graph data via the learning of an uncertainty-based hardness measure. To the best of our knowledge, it is the first work that addresses the problem using an uncertainty learning framework.
    \item We show theoretically that our approach transforms popular GCL losses such as InfoNCE into a triplet loss with an adaptive hardness-based margin, enforcing a large margin for hard negatives while pulling false negatives close to anchor instances.
    \item Extensive experiments on 10 graph datasets demonstrate the superiority of AUGCL in consistently enhancing different state-of-the-art GCL methods in both graph and node classification tasks (having maximal classification accuracy improvement by $\sim$2\% and $\sim$1.5\%, respectively), and the robustness against graph adversarial attacks (maximal improvement by $\sim$8\%).
\end{itemize}

\section{Related Works}
\label{relatedworks}

\subsection{Graph Contrastive Learning}
Recently, contrastive learning \cite{wu2018unsupervised, chen2020simple, van2018representation, he2020momentum} has become a prominent technique in self-supervised learning. It has been successfully adapted into diverse domains, including the graph domain. A number of GCL methods \cite{velickovic2019deep, hassani2020contrastive, sun2019infograph, tong2021directed, you2020graph, zhu2020deep, zhu2021graph, you2021graph, xia2022simgrace, xu2021infogcl, li2023afgcl} have been proposed. DGI \cite{velickovic2019deep} is an early attempt that obtained node representations by maximizing the mutual information between node embeddings and high-level graph information. MVGRL \cite{hassani2020contrastive} improved DGI by introducing different structural views to learn node and graph-level representations. InfoGraph \cite{sun2019infograph} performed contrastive learning by directly maximizing the consistency between sampled subgraphs and pooled graph representations. DiGCL \cite{tong2021directed} performed direct graph contrastive learning based on laplacian perturbation. Additionally, GraphCL \cite{you2020graph} systematically explored the influence of different augmentations on graph-level contrastive learning. GRACE \cite{zhu2020deep} designed a node-level contrastive framework by maximizing the agreement of node embeddings between two corrupted graph views. GCA \cite{zhu2021graph} conducted node-level contrastive learning with adaptive augmentation on the topology and node attribute level. G-Zoom \cite{zheng2022gzoom} learned node representation by establishing contrastiveness on different scales progressively. 
PA-GCL \cite{pagcl2023} explored channel-level contrastive learning among three generated proximity graphs and updated the generated graphs during the training process to improve their contrast effect.
Besides, some studies have been put forward to enhance the GCL by automating data augmentations \cite{you2021graph} or discarding explicit data augmentations \cite{xia2022simgrace} \cite{li2023afgcl}. InfoGCL \cite{xu2021infogcl} further built principles to build the optimal GCL model based on Information Bottleneck \cite{tishby2000information}. The main differences among these methods lie in the way they obtain positive pairs. By contrast, our approach AUGCL focuses on hard negative mining, which is orthogonal to these GCL methods and can be plugged into their loss function to improve their performance on graph/node-level tasks.

\subsection{Hard Negative Mining in Contrastive Learning}
Hard negative mining refers to generating or mining the negatives that are difficult to discriminate from the positive. Various methods have been proposed to perform hard negative mining to facilitate contrastive learning, including employing mixup strategy \cite{zhang2018mixup} to mix the anchor instance and negative instance to synthesize hard negatives \cite{kalantidis2020hard, verma2021towards, lee2021imix, ge2021robust}, and developing unsupervised sampling methods for selecting hard negative samples \cite{chuang2020debiased,robinson2020hard}. 
Specifically, DCL \cite{chuang2020debiased} adopted positive-unlabeled learning \cite{chen2020self} to avoid sampling false negatives. HCL \cite{robinson2020hard} improved DCL \cite{chuang2020debiased} by further up-weighting the negatives close to the anchor.
These methods are mainly focused on image data and they often treat negative instances that are most similar to the anchor instance as the hard negatives. However, for graph data, the similar negatives could be false negatives relative to the anchor, and the GCL performance would be degraded by employing these hard negative mining methods \cite{xia2022progcl, zhu2021empirical}. To address this issue, ProGCL \cite{xia2022progcl} exploited a two-component beta mixture model on the similarity distribution of negatives to estimate the probability of negative instances being true for an anchor and then measured the hardness of negative instances by integrating the estimated posterior probability and the similarity between the negative and the anchor.
AdaS \cite{adas2023} devised an adaptive sampling strategy to emphasize the negatives that are neither too hard nor too easy to discriminate from the anchor, and employed a polarization regularizer to enhance the influence of them.
HSAN \cite{HSAN} explored hard positives and negatives by introducing a weight-modulating function. Specifically, this function adjusts the weights of sample pairs in contrastive learning by calculating the disparity between high-confidence pseudo labels and the similarity obtained from both attribute and structure information. 
Instead of measuring the hardness of negatives, HCHSM \cite{tu2023hier} selected hard samples based on a mutual information agreement gap between positive and negative pairs and performed hierarchically contrastive learning for hard samples to exploit multilevel intrinsic graph features.
Similar to ProGCL, our method also measures the hardness of negatives for each anchor instance. However, we employ the uncertainty estimation model to directly learn the negative instance hardness. The learned hardness is then incorporated into the contrastive loss via a weighting term, resulting in an anchor-instance-adaptive contrastive learning framework with good theoretical support. 

\subsection{Uncertainty Estimation}
Numerous methods and theories have been introduced to measure the prediction uncertainty, e.g., by using the maximum of predicted probabilities \cite{geifman2017selective,lakshminarayanan2017simple, liang2018enhancing}, the prediction entropy/energy \cite{lakshminarayanan2017simple, gal2016dropout,liu2020energy,tian2021pixel}, or an extra (void/background) class \cite{lee2018training,liu2019deep,tian2021pixel}. These methods focus on calibrating prediction confidence in supervised learning, whereas we utilize uncertainty estimation under the self-supervised setting to empower contrastive learning. Our work is motivated by the observation that hard samples are typically the instances at the decision boundary between the positive and negative instances, which are also the samples that learning models are uncertain about. Thus, uncertainty estimation offers an effective way to measure the hardness of negative instances. To be applicable in graph contrastive learning, AUGCL is designed in a novel way by using an anchor-instance-dependent uncertainty learning approach.

\section{AUGCL: Affinity Uncertainty-based Graph Contrastive Learning}
\label{method}

\subsection{Preliminaries}
Self-supervised graph representation learning has demonstrated promising performance in empowering diverse graph learning tasks. This work focuses on node-level and graph-level tasks. Particularly, let $\mathcal{G} = (\mathcal{V}, \mathcal{E})$ denote a graph where $\mathcal{V}$ and $\mathcal{E}$ denote the set of nodes and edges respectively, then for a node-level task, the goal of self-supervised graph representation learning is to leverage a single graph $\mathcal{G}$ to learn an encoder $\psi(\mathcal{V}, \mathcal{E})$ without using the labels of nodes so that $\psi(\mathcal{V}, \mathcal{E})$ can yield an expressive low-dimensional embedding $z_i$ for each node in $\mathcal{V}$. The resulting node embeddings $\mathcal{Z} = \{z_i\}_{i=1}^{|\mathcal{V}|}$ can then be used in various downstream node-level tasks, such as node classification. For a graph-level task, the goal instead is to learn a graph encoder $\psi(\mathcal{V}_i, \mathcal{E}_i)$ given a set of $N$ graphs $\{\mathcal{G}_i=(\mathcal{V}_i, \mathcal{E}_i)\}_{i=1}^N$, where the encoder $\psi(\mathcal{V}_i, \mathcal{E}_i)$ outputs a low-dimensional embedding $z_i$ for each graph $\mathcal{G}_i$, and the graph embeddings $\mathcal{Z} = \{z_i\}_{i=1}^{N}$ can then be used in various downstream graph-level tasks, e.g., graph classification. Our approach can be used to improve the self-supervised learning of graph representations and node representations, as shown in Sec. \ref{Experiments}.
Without loss of generality, we use the graph-level tasks to introduce our approach below.

\subsection{Popular GCL Methods and Their Weaknesses}\label{weaknesses}

Graph contrastive learning is one of the most popular approaches for self-supervised graph representation learning. As an instance-wise discriminative approach, it aims to pull two different augmentations of the same graph closer and push augmentations of different graphs apart \cite{you2020graph, hassani2020contrastive}. InfoNCE \cite{van2018representation} is among the most popular contrastive learning loss functions to achieve this goal. Specifically, given a minibatch of randomly sampled graphs $\{\mathcal{G}_i\}_{i=1}^{N}$, two augmentation functions $t_1$ and $t_2$ are first sampled from the augmentation pool $\mathcal{T}$ which consists of all possible augmentations. Then, two graph views $\{\widetilde{\mathcal{G}}_i\}_{i=1}^{N}$ and $\{\widehat{\mathcal{G}}_i\}_{i=1}^{N}$ are generated by applying $t_1, t_2$ to each graph. The embeddings $\{\widetilde{z}_i\}_{i=1}^N$ and $\{\widehat{z}_i\}_{i=1}^N$ of the augmented graphs are obtained by feeding the augmented graphs into a shared GNN encoder $\psi(\cdot)$, followed by a projection head (2-layer perceptron) \cite{chen2020simple}. 
For an anchor instance $\widetilde{\mathcal{G}}_i$ -- a graph augmented from $\mathcal{G}_i$ using $t_1$, the positive is $\widehat{\mathcal{G}}_i$ -- a graph augmented from the same graph $\mathcal{G}_i$ but using a different augmentation $t_2$, while the source of the negative instances is $\{\widehat{\mathcal{G}}_j\}_{j=1}^{N}$, from which negative instances are sampled. To enforce the maximization of the consistency between positive embeddings, the pairwise objective for a positive pair $(\widetilde{z}_i, \widehat{z}_i)$ is formulated as:
\begin{equation}\label{infonce}
    \ell_{\text{InfoNCE}}(\widetilde{z}_i, \widehat{z}_i) = - \log \frac{e^{ (h(\widetilde{z}_i, \widehat{z}_i)/\tau)}}{e^{({h(\widetilde{z}_i, \widehat{z}_i)/\tau})}+ \sum\limits_{j, j\neq i}^N e^{({h(\widetilde{z}_i, \widehat{z}_j)/\tau})}},
\end{equation}
where $\tau$ denotes the temperature parameter and $h(\widetilde{z}_i, \widehat{z}_j)$ is the  cosine similarity function measuring similarity between $\widetilde{z}_i$ and $\widehat{z}_j$.

Although these graph contrastive learning methods have achieved great success in graph representation learning, they often fail to consider the semantics of negatives in $\{\widehat{\mathcal{G}}_j\}_{j=1}^{N}$. Consequently, instances that share the same semantics with the positive can be sampled and treated as negatives in (\ref{infonce}). This false negative sampling issue, also known as sampling bias in \cite{chuang2020debiased}, would hinder the learning of contrastive representations between positive instances and negative instances. More importantly, the contrastive learning cannot exploit \textit{hard negatives}, i.e., instances that are similar to but semantically different from the anchor, which are driving force for contrastive learning to learn substantially enhanced discriminative representations, as shown in the literature empirically and theoretically \cite{robinson2020hard, kalantidis2020hard, xia2022progcl}.

\begin{figure*}
    \centering
    \includegraphics[width=0.95\textwidth]{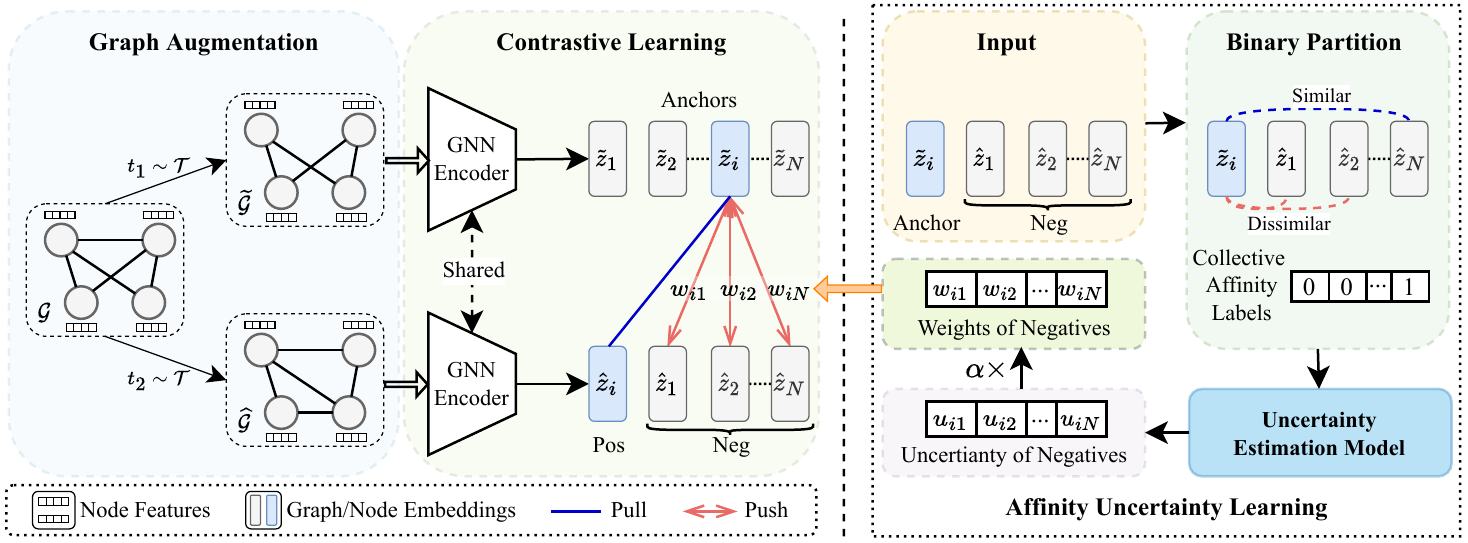}
    \caption{Overview of our approach AUGCL. \emph{Left}: AUGCL-based graph contrastive learning. The objective and the general procedures are the same as existing GCL methods, but AUGCL leverages affinity uncertainty to learn anchor-instance-dependent hardness-based instance weights $\{w_{i1},w_{i2},\cdots, w_{iN}\}$ for all negative instances to improve existing GCL methods. \emph{Right}: The proposed affinity uncertainty learning approach to obtain the weights. For an anchor $\widetilde{z}_i$, AUGCL first obtains collective affinity information (i.e., pairwise affinity across the instances) via binary partition of its negative instances. It then utilizes those affinity information to learn an uncertainty estimator that evaluates how confident the estimator is about the affinity of each negative instance $\widehat{z}_j$ relative to the anchor instance $\widetilde{z}_i$. A larger affinity uncertainty value $u_{ij}$ indicates more likely of $\widehat{z}_j$ being a hard negative, and thus, a larger weight $w_{ij}$ ($w_{ij}=\alpha u_{ij}$ where $\alpha$ is a hyperparameter). 
    }
    \label{fig:framework}
\end{figure*}

\subsection{The Proposed Approach AUGCL}
\label{sec:frameworkandtheory}

To address the negative sampling weaknesses discussed in Sec. \ref{weaknesses}, we present a novel framework for learning an affinity uncertainty-based hardness measure for enhancing current state-of-the-art graph contrastive learning methods. The key idea is to first learn the hardness of a negative instance relative to each anchor instance by comparing the affinity between them to the affinities of the anchor instance to the other instances. The hardness results can then be plugged into a contrastive loss, e.g., InfoNCE, to improve the effectiveness of current GCL methods in utilizing the hard negatives.

\textbf{Overview of AUGCL.} Since the hardness of a negative instance varies largely w.r.t. different anchor instances, our approach AUGCL aims to learn a hardness measure based on the relative affinity between the negative instance and each anchor instance. That is, for an anchor instance $\widetilde{z}_i$ and its negative instance candidate set $\mathcal{\widehat{Z}}_i=\{\widehat{z}_j\}_{j=1}^N$, we learn a single hardness measure function $\phi(\widehat{z}_j|\widetilde{z}_i;\Theta): \mathcal{\widehat{Z}}_i \rightarrow \mathbb{R}$ that yields a hardness value for each $\widehat{z} \in \mathcal{\widehat{Z}}_i $ relative to $\widetilde{z}_i$. Note that the function $\phi$ parameterized by $\Theta$ is trained across all anchor instances; yet the hardness it yields for the negative instance $\widehat{z}_j$ is dependent on the anchor $\widetilde{z}_i$. For brevity, $\phi(\widehat{z}_j|\widetilde{z}_i;\Theta)$ is denoted as $\phi_i(\widehat{z}_j;\Theta)$ hereafter.

Unlike current hardness measures that define the hardness of a negative instance based on its individual relation to the anchor instance (e.g., the similarity between them), one key novelty in AUGCL is that it defines the hardness based on two groups of pairwise affinities between the negative instances and the anchor instance. More specifically, we introduce the concept of affinity uncertainty below to achieve this goal:
\newtheorem{definition}{Definition}
\begin{definition}[Affinity Uncertainty]\label{def:au}
Given an anchor instance $\widetilde{z}_i$ and its negative instance candidate set $\mathcal{\widehat{Z}}_i = \{\widehat{z}_j\}_{j=1}^N$, and let $\mathcal{C}_1^i$ and $\mathcal{C}_2^i$ be two disjoint groups of instances in $\mathcal{\widehat{Z}}_i$ such that: one group $\mathcal{C}_1^i$ includes the instances that are closely aligned and distributed around the anchor $\widetilde{z}_i$, while the other group $\mathcal{C}_2^i$ contains the rest of other instances; and $\mathcal{\widehat{Z}}_i$= $\mathcal{C}_1^i \cup \mathcal{C}_2^i$. Then the affinity uncertainty of each $\widehat{z} \in \mathcal{\widehat{Z}}_i $ w.r.t. $\widetilde{z}_i$ is defined as:
\begin{equation}\label{au}
    \phi_i(\widehat{z}) = g(\widehat{z},\mathcal{C}_1^i,\mathcal{C}_2^i),
\end{equation}
where $g$ is an uncertainty estimator that evaluates how confident the estimator is about the affinity of $\widehat{z}$ to the instances in the anchor instance-centered group $\mathcal{C}_1^i$ compared to the other group $\mathcal{C}_2^i$.
\end{definition}

The affinity uncertainty in (\ref{au}) takes a holistic approach that considers diverse affinities of the negative instances within and across the two groups $\mathcal{C}_1^i$ and $\mathcal{C}_2^i$ to learn an accurate hardness for each negative instance $\widehat{z}$. As shown in the literature (e.g., \cite{liu2019deep}) and Fig. \ref{fig:uncervisual1}, instances that are ambiguous to distinguish are assigned to large uncertainty values. These instances typically have a poor affinity to both groups $\mathcal{C}_1^i$ and $\mathcal{C}_2^i$, such as those located on the boundary between the two groups. By contrast, if the instances are coherently aligned within $\mathcal{C}_1^i$ or $\mathcal{C}_2^i$, their uncertainty would be small. Thus, this type of uncertainty can be naturally used to define the hardness of the negative instances.

The obtained hardness can then be easily plugged into existing contrastive losses, such as the InfoNCE loss, via a weighting term for the negative instances. Particularly, the AUGCL-enhanced InfoNCE is given as follows: 
\begin{equation}\label{eqn:hardcl}
 \ell_{\text{AUGCL}}(\widetilde{z}_i, \widehat{z}_i) = 
- \log \frac{e^{ (h(\widetilde{z}_i, \widehat{z}_i)/\tau)}}{e^{({h(\widetilde{z}_i, \widehat{z}_i)/\tau})}+ \sum\limits_{j, j\neq i}^N w_{ij} e^{({h(\widetilde{z}_i, \widehat{z}_j)/\tau})}},
\end{equation}
where $w_{ij} = \alpha \phi_i(\widehat{z}_j; \Theta)$ is the hardness-based weight added to $\widehat{z}_j$ relative to $\widetilde{z}_i$. $\phi_i(\widehat{z}_j;\Theta)$ is the hardness learned by AUGCL for the negative instance $\widehat{z}_j$ w.r.t. the anchor instance $\widetilde{z}_i$ and $\alpha$ is a hyperparameter. This enables effective exploitation of the hard negatives, as large weights are expected for hard negatives while small weights are expected for the other instances, e.g., false negatives. 

The overall procedure of AUGCL is illustrated in Fig. \ref{fig:framework}. It follows the standard graph contrastive learning in the graph augmentation and contrastive learning except that we incorporate the affinity uncertainty-based hardness through a weighting term into the contrastive loss as in (\ref{eqn:hardcl}). The right panel in Fig. \ref{fig:framework} shows the steps of learning an anchor-dependent hardness measure $\phi$ for each anchor $\widetilde{z}_i$, consisting of instance partition and uncertainty estimation as indicated in Def. \ref{def:au}. Before introducing the details of these two components in Sec. \ref{uncertaintymodel1}, below we demonstrate the theoretical motivation of the proposed method.

\textbf{Theoretical Motivation.}
We show below that \eqref{eqn:hardcl} is equivalent to a triplet loss with an adaptive margin exponentially proportional to the learned hardness-based weight $\phi_i(\widehat{z}_j; \Theta)$. This provides a more straightforward explanation of the working mechanism of the proposed weighting method. 
\newtheorem{theorem}{Theorem}
\begin{restatable}{theorem}{resta}
\label{theorem}
Let $u_{ij}=\phi_i(\widehat{z}_j;\Theta)$ be the affinity uncertainty-based hardness of a negative instance $\widehat{z}_j$ w.r.t. the anchor instance $\widetilde{z}_i$. When the projection function is an identity function and assumes the positive instance is more similar to the anchor than the negative instances, then minimizing the proposed objective in \eqref{eqn:hardcl} is equivalent to minimizing a modified triplet loss with an adaptive margin $m_{ij} = \frac{\tau}{2} \log (\alpha u_{ij})$, i.e.,  
\begin{equation}
\label{hardcl_obj}
 \ell_{\text{AUGCL}}(\widetilde{z}_i, \widehat{z}_i) \propto \ \  \frac{1}{2\tau}\sum_{j, j\neq i}^{N}\left(\|\widetilde{z}_i^{'}- \widehat{z}_i^{'}\|- \|\widetilde{z}_i^{'} - \widehat{z}_j^{'}\| +m_{ij}\right),
\end{equation}
where $\widetilde{z}_i^{'}$ is the normalized embedding.
\end{restatable}

The proof for this theorem is detailed in the Appendix. From the theorem, we can see that the optimal embeddings to \eqref{hardcl_obj} should satisfy the following inequality:
\begin{equation}\label{inequation}
    \|\widetilde{z}_i^{'}- \widehat{z}_i^{'}\|  \ll \|\widetilde{z}_i^{'} - \widehat{z}_j^{'}\| - m_{ij},
\end{equation}
where $m_{ij} = \frac{\tau}{2} \log (\alpha u_{ij})$ and  $u_{ij}=\phi_i(\widehat{z}_j;\Theta)$. Thus, $m_{ij}$ is equivalent to a transformed affinity uncertainty-based hardness of the negative instance $\widehat{z}_j$ relative to the anchor $\widetilde{z}_i$, satisfying:
\begin{equation}
\begin{cases}
m_{ij} \geq 0, & \text{if $\alpha u_{ij} \geq 1$}; \\
m_{ij} < 0,    & \text{otherwise}.
\end{cases}
\end{equation}

If $\widehat{z}_j$ is a hard negative for $\widetilde{z}_i$, the large uncertainty $u_{ij}$ between $\widetilde{z}_i$ and $\widehat{z}_j$ makes the inequality \eqref{inequation} hard to satisfy through $m_{ij} > 0$, enforcing better representation learning. On the contrary, if the uncertainty $u_{ij}$ is small, \eqref{inequation} can be easily satisfied with $m_{ij} \ll 0$, reducing the impact of the possible false negative instances.

\subsection{Instantiation of AUGCL}
\label{uncertaintymodel1}
We introduce an instantiation of our AUGCL framework in this subsection.
As demonstrated in Def. \ref{def:au}, the affinity uncertainty-based hardness function $\phi$ parameterized with $\Theta$ can be decomposed into two modules, including a binary clustering function $f:\{\widehat{z}_j\}_{j=1}^N \rightarrow \{0, 1\}$ parameterized by $\Theta_f$ and an uncertainty estimation function $g:\{\widehat{z}_j\}_{j=1}^N \times  \{0, 1\} \rightarrow \mathbb{R}$ parameterized by $\Theta_g$, i.e., $\Theta=\{\Theta_f,\Theta_g\}$. AUGCL is a generic framework. Different clustering and uncertainty estimation methods can be adopted in AUGCL to implement a specific model, as shown by our empirical results in Sec. \ref{ablationstudy}. Below we describe the two modules of the best instantiated AUGCL model based on our experiments.

\subsubsection{Anchor-dependent Binary Partition of Negatives}

Given an anchor $\widetilde{z}_i$, binary clustering is used to partition the negative samples into two coherent groups -- $\mathcal{C}_1^i$ and $\mathcal{C}_2^i$ -- for subsequent affinity uncertainty estimation. Without having access to label information, clustering is often adopted on the full dataset to divide instances into several clusters \cite{caron2020unsupervised, li2020prototypical,lin2021prototypical, zhao2021graph, xu2021self, li2022graph, luo2022clear}, and instances from clusters other than the anchor-based cluster are directly treated as negatives. 

Our clustering differs from these existing methods in two main ways. First, we perform an anchor-dependent binary partition on only the negative instances in each batch of instances rather than the full dataset. Specifically, given a batch of node/graph embeddings $\{\widehat{z}_j\}_{j=1}^N$, for each anchor $\widetilde{z}_i \in \{\widetilde{z}_i\}_{i=1}^N$, we perform a binary partition on the negative instance candidates $\{\widehat{z}_j\}_{j=1}^N$ using an existing clustering method (e.g., $k$-means), i.e., $f_{k-\text{means}}: \{\widehat{z}_j\}_{j=1}^N \rightarrow \{0,1\}$.
As a result, the negative instances are partitioned into two clusters. We denote the cluster distributed around the anchor instance $\widetilde{z}_i$ as $\mathcal{C}_1^{i}$ and assign the label of 1 to the negative instances within $\mathcal{C}_1^{i}$, i.e., $C_1^{ij}=1$ if $\widehat{z_j} \in \mathcal{C}_1^{i}$. Simultaneously, we refer to the other cluster as $\mathcal{C}_2^{i}$ and assign the label of 0 to the negative instances within it.

The second difference is that the obtained partitions are used to gain a sense of the affinity of an instance to the other instances, rather than being the direct negative sample clusters. The affinity information would be used to evaluate the hardness of each negative instance through an uncertainty estimation model in AUGCL. 

\subsubsection{Affinity Uncertainty Estimation}

For an anchor $\widetilde{z}_i$, the binary cluster labels $\{C_{1|2}^{ij}\}_{j=1}^N$ carry the affinity semantics of the instances $\{\widehat{z}_j\}_{j=1}^N$  w.r.t. the anchor instance $\widetilde{z}_i$. We further propose to perform an uncertainty estimation upon these affinity semantic-based labels for each anchor $\widetilde{z}_i \in \{\widetilde{z}_i\}_{i=1}^N$, and use this uncertainty to measure the hardness of instances $\{\widehat{z}_j\}_{j=1}^N$. By doing so, a large uncertainty-based hardness is assigned to fringe instances that are located around the boundary between the two clusters; these instances are typically hard negatives w.r.t. $\widetilde{z}_i$. A small hardness is assigned otherwise. 

Different uncertainty estimation methods can be used to specify this component. We found that the recently proposed method Deep Gambler (DG) \cite{liu2019deep} worked best in our experiments, so DG is used in AUGCL by default. Specifically, DG extends a multi-class classification task to a problem that learns an extra class to represent the uncertainty of instances, in addition to guaranteeing the classification of the original classes. For an anchor instance $\widetilde{z}_i$, given its associated negative instance candidates $\{\widehat{z_j}\}_{j=1}^N$ and their affinity labels $\{C_{1|2}^{ij}\}_{j=1}^N$, the DG-based uncertainty estimation is trained by minimizing the following loss:
\begin{equation}\label{eqn:dg}
    \mathcal{\ell}_i^{DG} = -\sum_j^N \log(p_{C_{1|2}^{ij}} * o + u_{ij}),
\end{equation}
where $p_{C_{1|2}^{ij}}$ is the predicted class probability on class $C_{1|2}^{ij}$ from a multi-layer perceptrons-based (MLP-based) DG model $g(\widehat{z},\mathcal{C}_1^i,\mathcal{C}_2^i; \Theta_g)$ parameterized by $\Theta_g$, $u_{ij}$ is the uncertainty that the model $g$ generates for the instance $\widehat{z}_j$. $o$ is the reward parameter which is an important and necessary parameter for DG. A larger $o$ encourages DG to be more confident in inferring, i.e., producing a lower uncertainty value, and vice versa. The final loss of DG is computed across all anchor instances $\widetilde{z}_i \in \{\widetilde{z}_i\}_{i=1}^N$.

After the DG model is trained, for each anchor $\widetilde{z}_i$, we calculate $u_{ij}$ for its negative instance $\widehat{z}_j$ and obtain a uncertainty matrix $\mathbf{U} \in \mathbb{R}^{N\times (N-1)}$ where each row $\mathbf{u}_i$ contains the uncertainty of all negative instances w.r.t. the anchor $\widetilde{z}_i$. These uncertainty values are then used in \eqref{eqn:hardcl} to improve contrastive learning.

\subsection{Time Complexity Analysis}
We take $k$-means and Deep Gambler \cite{liu2019deep} as the partition and uncertainty estimation methods, respectively, to analyze the additional time complexity introduced by AUGCL. Specifically, let $L$ be the number of MLP layers in DG and $d$ be the number of hidden units for all layers. For the graph classification task, given a graph dataset with $N$ graphs and the batch size is set as $B$, the time complexities of partition and the uncertainty modeling are $\mathcal{O}(2(\frac{N}{B})B^2T)$ and $\mathcal{O}(KL(\frac{N}{B})B^2d^2)$ respectively, where $T$ is the number of iterations for $k$-means and $K$ is the number of training epochs for the uncertainty estimation model. For the node classification task, given a graph with $N$ nodes, in order to reduce the computation cost, we only sample $M$ ($M\ll N$) negatives for an anchor when training AUGCL. The resulting time complexities of the partition and training uncertainty model are $\mathcal{O}(2NMT)$ and $\mathcal{O}(KLNMd^2)$ respectively. In experiments, we use the well-established $k$-means clustering implementation from scikit-learn \cite{scikit-learn}, as it runs very fast in practice. Besides, the values of $K$, $L$, $M$ and $d$ are relatively small and the uncertainty estimation model is only trained once. Therefore, the computational overhead over the base model is not significant.

\section{Experiments}
\label{Experiments}

\subsection{Experimental Setup}

\subsubsection{Datasets}

Seven commonly used graph classification datasets are used in our experiments. They come from two popular application domains: bioinformatics (MUTAG, DD, NCI1, and PROTEINS) and social networks (COLLAB, REDDIT-BINARY, and IMDB-BINARY). For the node classification task, we use three widely used datasets, i.e., Wiki-CS \cite{mernyei2020wiki}, Amazon-Computers, and Amazon-Photo \cite{shchur2018pitfalls}. Wiki-CS is a reference network constructed based on Wikipedia. Amazon-Computers and Amazon-Photo are two co-purchase networks constructed from Amazon. The statistics of the datasets are summarized in Table~\ref{datasetall}.

\begin{table}[!hbtp]
\caption{Key statistics of datasets used.}
  \label{datasetall}
  \centering
  \resizebox{0.48\textwidth}{!}{
\begin{tabular}{c|cccc}
\hline
\textbf{Task}    & \textbf{Dataset} & \textbf{ Graphs} & \textbf{  Avg.Nodes} & \textbf{  Avg.Edges}  \\ \hline
\multirow{8}{*}{\begin{tabular}[c]{@{}c@{}}Graph\\ Classification\end{tabular}} & NCI1 & 4,110 & 29.87 &32.30\\
                       & PROTEINS & 1,113 & 39.06 & 72.82\\
                       & DD  & 1,178 & 284.32 &715.66\\
                       & MUTAG  & 188 & 17.93 & 19.79\\
                       & COLLAB  & 5,000 & 74.49 & 2,457.78 \\
                       & RDT-B  & 2,000 & 429.63 & 497.75 \\
                       & IMDB-B  & 1,000 & 19.77 & 96.53\\ \hline
\textbf{Task}  & \textbf{Dataset} & \textbf{ Graphs} & \textbf{  Nodes} & \textbf{  Edges}\\ \hline
\multirow{3}{*}{\begin{tabular}[c]{@{}c@{}}Node\\ Classification\end{tabular}}                        & Wiki-CS & 1  & 11,701 &216,123  \\ 
                       & Aamazon-Computers & 1 & 13,752  & 245,861 \\ 
                       & Aamazon-Photo & 1 & 7,650  & 119,081 \\ \hline
\end{tabular}
}
\end{table}

\subsubsection{Implementation Details and Evaluation Protocol}

For the graph classification task, GraphCL \cite{you2020graph}, a recent SOTA InfoNCE-based contrastive learning method for graph classification, is used as our base, into which our affinity uncertainty-based hardness learning method is plugged. For a fair comparison, the network backbone, the graph augmentation methods and the hyper-parameters of our AUGCL-enabled GraphCL are kept exactly the same as the original GraphCL. We follow a widely-used two-stage evaluation protocol in the literature \cite{sun2019infograph, you2020graph, yanardag2015deep, narayanan2017graph2vec}, in which we first learn graph representations in a self-supervised manner and then use the representations to train a downstream SVM classifier. The 10-fold evaluation is adopted in classification, and it is repeated five times with the mean accuracy (\%) and standard variation reported.

For the node classification task, we adopt GCA \cite{zhu2021graph} as the base model and plug our AUGCL-based affinity uncertainty hardness into it. The evaluation protocol for node classification follows DGI \cite{velickovic2019deep} where the model is first trained in an unsupervised manner and then the learned node representations are used to train and test a simple $\ell_2$-regularized logistic regression classifier. On each dataset, the experiment is repeated for 20 runs with different data splits, and the average classification accuracy, together with the standard variation, is reported.

For graph and node classification, we use the same architecture in our affinity uncertainty estimation model, i.e., a three-layer multi-layer-perceptrons (MLP) architecture, containing 128 units per layer with $ReLU$ activation. We adopt the Stochastic Gradient Descent (SGD) optimizer for the uncertainty estimation model and the learning rate is set to 0.01 across all the datasets. The uncertainty scaling parameter $\alpha$ is set to the reciprocal of the mean of uncertainties. The training epoch number of the uncertainty estimation model is set to 10 for all datasets. For the reward parameter in the uncertainty estimation model, it is selected through a grid search, and the search space is $\{1.5, 1.6, 1.7, 1.8, 1.9\}$.

\begin{table}[htbp]
  \caption{Comparison between the baselines and their AUGCL-enabled counterparts. The baselines are GraphCL \cite{you2020graph} and GCA \cite{zhu2021graph} for graph and node classification tasks, respectively. ``$\textcolor{magenta}\uparrow$'' refers to the improvement compared to the baselines.}
  \label{comp_base_our}
  \centering
  \resizebox{0.48\textwidth}{!}{
  \begin{tabular}{c|cccc}
    % \toprule
    \hline\hline
    \textbf{Task} & \textbf{Dataset} & \textbf{GraphCL} &\textbf{GraphCL$_{\text{AUGCL}}$}\\
    % \midrule
    \hline
    \multirow{7}{*}{\begin{tabular}[c]{@{}c@{}}Graph\\ Classification\end{tabular}}
    & NCI1 &78.26 &80.16($\textcolor{magenta}\uparrow$ 1.90) \\
    & PROTEINS &74.36 &75.76($\textcolor{magenta}\uparrow$ 1.40) \\
    & DD &78.01 &80.14 ($\textcolor{magenta}\uparrow$ 2.13) \\
    & MUTAG &87.15 &89.20 ($\textcolor{magenta}\uparrow$ 2.05) \\
    & COLLAB &71.53 &72.10 ($\textcolor{magenta}\uparrow$ 0.57)\\
    & RDT-B &90.09 &91.19 ($\textcolor{magenta}\uparrow$ 1.10) \\
    & IMDB-B &71.20 &72.46 ($\textcolor{magenta}\uparrow$ 1.26) \\
    \hline
    \textbf{Task} & \textbf{Dataset} & \textbf{GCA} &\textbf{GCA$_{\text{AUGCL}}$}\\\hline
    \multirow{3}{*}{\begin{tabular}[c]{@{}c@{}}Node\\ Classification\end{tabular}}
    & Wiki-CS &78.08 &78.59 ($\textcolor{magenta}\uparrow$ 0.51) \\
    & Amazon-Computers&87.80 &88.94 ($\textcolor{magenta}\uparrow$ 1.14) \\
    & Amazon-Photo&91.99 &93.43 ($\textcolor{magenta}\uparrow$ 1.44) \\
    \hline    \hline
  \end{tabular}
}
\end{table}

\begin{table*}[htbp]
  \caption{Results of graph classification accuracy (\%). We obtain the results of GraphCL and its four hardness-aware variants under the same experimental setting as \cite{you2020graph}, from which the results of the other methods are taken. The results of JOAOv2 \cite{you2021graph} and SimGRACE \cite{xia2022simgrace} are taken from their own paper.}
  \label{results}
  \centering
  \resizebox{0.95\textwidth}{!}{
  \begin{tabular}{c|l|ccccccc}
    % \toprule
    \hline\hline
    \textbf{Type} & \textbf{Method} & \textbf{NCI1} & \textbf{PROTEINS} & \textbf{DD} & \textbf{MUTAG} & \textbf{COLLAB} & \textbf{RDT-B} & \textbf{IMDB-B}\\
    % \midrule
    \hline
    \multirow{6}{*}{\begin{tabular}[c]{@{}c@{}} Non-Contrastive \\ Methods \end{tabular}}
    & GK &-   &-& -& 81.66$\pm$2.11& - &77.34$\pm$0.18& 65.87$\pm$0.98\\
    & WL &80.01$\pm$0.50& 72.92$\pm$0.56& -& 80.72$\pm$3.00& - &68.82$\pm$0.41& 72.30$\pm$3.44\\
    & DGK &\textbf{80.31$\pm$0.46}& 73.30$\pm$0.82& - &87.44$\pm$2.72& - &78.04$\pm$0.39& 66.96$\pm$0.56\\
    & node2vec &54.89$\pm$1.61& 57.49$\pm$3.57& - &72.63$\pm$10.20& -& -& -\\
    & sub2vec  &52.84$\pm$1.47& 53.03$\pm$5.55& - &61.05$\pm$15.80& -& 71.48$\pm$0.41& 55.26$\pm$1.54\\
    & graph2vec&73.22$\pm$1.81& 73.30$\pm$2.05& -& 83.15$\pm$9.25& - &75.78$\pm$1.03 & 71.10$\pm$0.54\\
    \hline
    \multirow{4}{*}{\begin{tabular}[c]{@{}c@{}} Contrastive \\ Methods \end{tabular}}
    & InfoGraph&76.20$\pm$1.06&74.44$\pm$0.31& 72.85$\pm$1.78& 89.01$\pm$1.13& 70.65$\pm$1.13& 82.50$\pm$1.42&  \textbf{73.03$\pm$0.87}\\
    & JOAOv2 & 78.36$\pm$0.53 & 74.07$\pm$1.10 & 77.40$\pm$1.15 & 87.67$\pm$0.79 & 69.33$\pm$0.34 & 86.42$\pm$1.45 & 70.83$\pm$0.25\\
    &SimGRACE & 79.12$\pm$0.44 & 75.35$\pm$0.09 & 77.44$\pm$1.11 & 89.01$\pm$1.31 & 71.72$\pm$0.82 & 89.51$\pm$0.89 & 71.30$\pm$0.77\\
    & GraphCL & 78.26$\pm$0.39& 74.36$\pm$0.44 & 78.01$\pm$0.77 &87.15$\pm$0.86 & 71.53$\pm$1.03 &90.09$\pm$0.70 &71.20$\pm$0.72 \\
    \hline
    \multirow{4}{*}{\begin{tabular}[c]{@{}c@{}} Hardness-aware \\ Methods \end{tabular}}
    & GraphCL$_{\text{DCL}}$ & 77.62$\pm$0.67	&74.73$\pm$0.39	&76.84$\pm$1.24	&88.28$\pm$1.75	&70.36$\pm$0.97	&89.88$\pm$0.72& 70.62$\pm$0.58\\
    & GraphCL$_{\text{HCL}}$ &78.16$\pm$0.53&	74.39$\pm$0.77&	76.83$\pm$1.15&	88.94$\pm$1.22&	70.37$\pm$0.33&	90.05$\pm$0.47&	71.38$\pm$0.62\\
    & GraphCL$_{\text{ProGCL}}$ &76.93$\pm$0.47 &74.48$\pm$0.37 &79.22$\pm$0.90 &88.73$\pm$1.40&70.46$\pm$0.28 & 90.51$\pm$0.49 &71.58$\pm$0.59 \\
    & GraphCL$_{\text{AUGCL}}$ (Ours) &80.16$\pm$0.34& \textbf{75.76$\pm$0.39}&	\textbf{80.14$\pm$0.54}&	\textbf{89.20$\pm$1.08}& \textbf{72.10$\pm$0.65}& \textbf{91.19$\pm$0.44}&72.46$\pm$0.80\\
    \hline\hline
  \end{tabular}
}
\end{table*}

\begin{table}[htbp]
  \caption{Results of node classification accuracy (\%). We obtain the results of GCA and its four hardness-aware variants under the same experimental setting as \cite{zhu2021graph}, from which the results of the other methods are taken.}
  \label{noderesults1}
  \centering
  \resizebox{0.50\textwidth}{!}{
  \begin{tabular}{c|l|ccc}
    % \toprule
    \hline\hline
    \textbf{Type} & \textbf{Method} & \textbf{Wiki-CS} & \textbf{\begin{tabular}[c]{@{}c@{}} Amazon- \\ Computers \end{tabular}} & \textbf{\begin{tabular}[c]{@{}c@{}} Amazon- \\ Photo \end{tabular}}\\
    % \midrule
    \hline
    \multirow{6}{*}{\begin{tabular}[c]{@{}c@{}} Non-Contrastive \\ Methods \end{tabular}}
    & Raw feature &71.98$\pm$0.00 &73.81$\pm$0.00 &78.53$\pm$0.00\\
    & node2vec &71.79$\pm$0.05 &84.39$\pm$0.08 &89.67$\pm$0.12\\
    & DW &74.35$\pm$0.06 &85.68$\pm$0.06& 89.44$\pm$0.11 \\
    & DW+feature & 77.21$\pm$0.03& 86.28$\pm$0.07& 90.05$\pm$0.08\\
    & GAE &70.15$\pm$0.01& 85.27$\pm$0.19&91.62$\pm$0.13\\
    & VGAE & 75.63$\pm$0.19& 86.37$\pm$0.21& 92.20$\pm$0.11\\
    \hline
    \multirow{4}{*}{\begin{tabular}[c]{@{}c@{}} Contrastive \\ Methods \end{tabular}}
    & DGI & 75.35$\pm$0.14& 83.95$\pm$0.47& 91.61$\pm$0.22\\
    & GMI &74.85$\pm$0.08& 82.21$\pm$0.31 &90.68$\pm$0.17\\
    & MVGRL&77.52$\pm$0.08& 87.52$\pm$0.11 &91.74$\pm$0.07\\
    & GCA &78.08$\pm$0.58 & 87.80$\pm$0.42 &91.99$\pm$0.39 \\
    \hline
    \multirow{4}{*}{\begin{tabular}[c]{@{}c@{}} Hardness-aware \\ Methods \end{tabular}}
    & HSAN & 78.55$\pm$0.51 & 81.23$\pm$1.05 & 89.15$\pm$0.66  \\
    & GCA$_{\text{DCL}}$ &78.12$\pm$0.60 &86.79$\pm$0.48 &91.29$\pm$0.32 \\
    & GCA$_{\text{HCL}}$ &78.19$\pm$0.64 &87.64$\pm$0.34 & 91.79$\pm$0.29\\
    & GCA$_{\text{ProGCL}}$ &78.33$\pm$0.64 &88.68$\pm$0.35 &93.01$\pm$0.29 \\
    % \hline
    % \cline{2-5}
    & GCA$_{\text{AUGCL}}$ (Ours) &\textbf{78.59$\pm$0.56} &\textbf{88.94$\pm$0.44} &\textbf{93.43$\pm$0.32} \\
    \hline\hline
  \end{tabular}
}
\end{table}

\subsubsection{Competing Methods}
We evaluate the effectiveness of AUGCL in both graph and node classification tasks. In both tasks, AUGCL is evaluated against three state-of-the-art hard negative mining-based contrastive learning methods, including DCL \cite{chuang2020debiased}, HCL \cite{robinson2020hard} and ProGCL \cite{xia2022progcl}. For the node classification task, we further compare the proposed method with the recently proposed hard sample aware method HSAN \cite{HSAN}. While HSAN was initially designed for node clustering, we adapt it to node classification by replacing the clustering operation with the node-level evaluation protocol mentioned above.

For both tasks, we also include a set of other relevant state-of-the-art competing methods, including non-contrastive methods and other contrastive methods. Particularly, for graph classification, the non-contrastive methods include Graphlet Kernel (GK) \cite{shervashidze2009efficient}, Weisfeiler-Lehman Sub-tree Kernel (WL) \cite{shervashidze2011weisfeiler}, Deep Graph Kernels (DGK) \cite{yanardag2015deep}, node2vec \cite{grover2016node2vec}, sub2vec \cite{adhikari2018sub2vec} and graph2vec\cite{narayanan2017graph2vec}, while the GCL methods include InfoGraph \cite{sun2019infograph}, JOAOv2 \cite{you2021graph}, SimGRACE \cite{xia2022simgrace} and GraphCL \cite{you2020graph}. For the node classification task, non-contrastive methods include node2vec \cite{grover2016node2vec}, DeepWalk (DW) \cite{perozzi2014deepwalk}, and Graph AutoEncoders (GAE and VGAE) \cite{kipf2016variational}. Contrastive methods include DGI \cite{velickovic2019deep}, GMI \cite{peng2020graph}, MVGRL \cite{hassani2020contrastive}, and GCA \cite{zhu2021graph}. 

Note that ProGCL proposed two strategies to utilize the estimated hardness results, i.e., weighting and mixup. The results reported are based on the weighting strategy of ProGCL to have a direct comparison to our weighting-based AUGCL.

\begin{table*}[!htbp]
    \caption{Classification accuracy under three evasion attacks on three different layers of structure2vec \cite{dai2016discriminative}. ProGCL and AUGCL below represent GraphCL methods with their respective hard negative mining component.}
    \label{adr}
    \centering
    \resizebox{1.0\textwidth}{!}{
    \begin{tabular}{l|cccc|cccc|cccc}
    \hline \hline
    % \hline
    \multirow{2}{*}{ Attacks } & \multicolumn{4}{|c|}{ 2-Layer } & \multicolumn{4}{|c|}{ 3-Layer } & \multicolumn{4}{c}{ 4-Layer } \\
    \cline { 2 - 13 } & Original & GraphCL & ProGCL & AUGCL & Original & GraphCL  & ProGCL & AUGCL & Original & GraphCL & ProGCL & AUGCL\\
    % \hline 
    \hline 
    Unattack & $93.20$ & {\bf 94.73} & 94.13 & 94.20 & $98.20$ & $98.33$ & 98.67 & {\bf 98.87} & $98.87$ & $99.00$ & {\bf 99.47} & 99.20 \\
    \hline 
    RandSampling & $78.73$ & $80.68$ & 82.47 & {\bf 82.67} & $92.27$ & $92.60$ & 93.93 & {\bf 94.67} & $95.13$ & $97.40$ & 97.13 & {\bf 97.93}\\
    GradArgmax & $69.47$ & $69.26$ & 74.80 & {\bf 77.53} & $64.60$ & $89.33$ & {\bf 94.07} & 93.27 & $95.80$ & \textbf{97.00} & 95.67 & 96.47\\
    RL-S2V & \textbf{42.93} & $42.20$ & 42.13 & 42.47 & $41.93$ & $61.66$ & 62.07 & {\bf 63.73} & $70.20$ & $84.86$ & 86.67 & \textbf{87.33} \\
    \hline  \hline
    \end{tabular}
    }
\end{table*}

\begin{table*}[htbp]
  \caption{Ablation study results of GraphCL$_{\text{AUGCL}}$ using different clustering methods $f$ or uncertainty estimation methods $g$.
    }\label{ablation_results}
  \centering
  \resizebox{1.0\textwidth}{!}{
  \begin{tabular}{l|cc|ccccccc}
    \hline\hline
    \textbf{Method} & $f$ & $g$ & \textbf{NCI1} & \textbf{PROTEINS} & \textbf{DD} & \textbf{MUTAG} & \textbf{COLLAB} & \textbf{RDT-B} & \textbf{IMDB-B}\\
    \hline
    \hline
    GraphCL (Baseline) & $\times$&$\times$ & 78.26$\pm$0.39& 74.36$\pm$0.44 & 78.01$\pm$0.77 &87.15$\pm$0.86 & 71.53$\pm$1.03 &90.09$\pm$0.70 &71.20$\pm$0.72 \\\hline
    \multirow{5}{*}{GraphCL$_{\text{AUGCL}}$}
    & Spectral & ExtraClass & 79.79$\pm$0.52 & 75.57$\pm$0.79 & 79.05$\pm$0.44 & 88.86$\pm$2.22 & 71.74$\pm$0.88	& 91.01$\pm$0.56 & 71.86$\pm$0.47 \\
     & $k$-means & ExtraClass &\textbf{80.16$\pm$0.34}& \textbf{75.76$\pm$0.39}&	\textbf{80.14$\pm$0.54}&	\textbf{89.20$\pm$1.08}& \textbf{72.10$\pm$0.65}& \textbf{91.19$\pm$0.44}& \textbf{72.46$\pm$0.80}\\
    \cline{2-10}
     & $k$-means & Distance & 78.88$\pm$0.26 &75.48$\pm$1.06  & 79.22$\pm$0.25 &88.64$\pm$0.83 	& 71.27$\pm$0.43 &90.41$\pm$0.51 & 72.20$\pm$0.55\\
     & $k$-means& Softmax & 79.42$\pm$0.43 & 75.34$\pm$0.57 & 78.10$\pm$0.65 & 86.81$\pm$1.65 &71.89$\pm$0.92 & 90.38$\pm$0.33  & 71.72$\pm$0.57\\
     & $k$-means& Entropy & 79.98$\pm$0.50 & 75.13$\pm$0.39 & 79.58$\pm$0.49 & 88.98$\pm$1.78	& 71.71$\pm$0.37 & 90.37$\pm$0.50  & 71.76$\pm$0.57\\
    \hline
    % \hline
  \end{tabular}
  }
\end{table*}

\subsection{Enabling Different GCL Methods on Graph and Node Classification}\label{sec:graph_node_results}

\subsubsection{Performance Improvement over Baselines.}
We first compare the performance of our method with the baselines on graph and node classification tasks. The results are shown in Table~\ref{comp_base_our}. It is clear that, by incorporating our affinity uncertainty-based hardness measure, the two baselines -- GraphCL \cite{you2020graph} and GCA \cite{zhu2021graph} -- are substantially and consistently boosted on all datasets from different domains for both graph and node classification tasks. This demonstrates that our method AUGCL can enable these baselines to effectively attend to hard negative instances and learn better representations of graphs/nodes.

\subsubsection{Comparison to State-of-the-art Methods} 
We then compare AUGCL to diverse advanced graph embedding learning methods.

\textbf{Graph Classification.} The results on graph classification are reported in Table~\ref{results}. We can observe that graph contrastive learning methods generally obtain better performance than non-contrastive methods. Our method further improves the performance by learning and feeding the affinity uncertainty-based hardness into contrastive learning, substantially outperforming SOTA GCL methods on 6 out of 7 datasets.

Compared to the three recent hardness-aware methods DCL, HCL and ProGCL, our method AUGCL performs much better across all seven datasets. Particularly, DCL, HCL and ProGCL improve GraphCL on some datasets such as PROTEINS, MUTAG, and IMDB-B, but they fail on the other ones. By contrast, our method improves over GraphCL by a large margin across all the seven datasets, indicating the superiority of our affinity uncertainty-based hardness learning method over its recent counterparts.

\textbf{Node Classification.}
The node classification results are reported in Table~\ref{noderesults1}. In general, the trends here are similar to the results in Table~\ref{results}: i) contrastive methods are generally more effective than the non-contrastive ones, ii) the competing hardness-aware methods DCL, HCL and ProGCL further improve over the contrastive methods on part of the datasets, while our method AUGCL achieves consistently better performance on all the three datasets, and iii) although HSAN \cite{HSAN} showed impressive performance in enabling node clustering tasks compared to GCA \cite{zhu2021graph}, it does not demonstrate consistently promising performance in our evaluation tasks. This discrepancy may arise from the fact that HSAN and GCA use different backbones and HSAN was not originally designed for node classification tasks.

From the results on graph and node classification tasks, we can see that 
general hardness-aware methods DCL and HCL do not achieve satisfactory performance on both tasks. This can be attributed to that they treat negatives similar to the anchor as hard negatives. Different from independent data, the non-i.i.d nature of graph data and the message-passing mechanism of graph neural networks result in similar negatives having the same label as the anchor with high probability, making DCL and HCL less effective for graph contrastive learning. To address this issue, ProGCL measured the hardness of negatives based on the similarity between negatives and the anchor and the learned probability of a negative being true from a two-component beta mixture model.
However, the prior employed by ProGCL that the negative similarity distribution is bimodal limits the applicability of ProGCL. Thus, ProGCL is not generalizable to the graph classification task, i.e., ProGCL fails to work as effectively as the baseline GraphCL on some datasets as shown in Table~\ref{results}, while our method AUGCL consistently outperforms the baseline. This is due to that our method learns a data-driven affinity uncertainty estimation model without the prior assumption, resulting in better applicability and flexibility of AUGCL on different graph mining tasks than ProGCL.

\subsection{Improving Robustness against Graph Adversarial Attacks} 

Self-supervised learning has shown effective performance in defending against adversarial perturbations \cite{hendrycks2019using,kim2020adversarial}. This subsection investigates whether AUGCL can further improve over the GCL methods on this important property. In this experiment, following \cite{dai2018adversarial}, three different types of graph adversarial attacks: RandSampling, GradArgmax and RL-S2V are used, where RandSampling randomly adds or deletes edges from graphs, GradArgmax performs edge modification based on gradient information, and RL-S2V is a reinforcement learning based attack method that learns a generalizable attack policy. We also use the widely-used evaluation protocol as in \cite{dai2018adversarial} where the graph task is to classify the component numbers in synthetic graphs and structure2vec \cite{dai2016discriminative} is adopted as the graph encoder, with different depths of structure2vec considered in the experiments. Both the original structure2vec trained from scratch (i.e., no pre-training) and the pre-trained structure2vec \cite{dai2016discriminative} using GraphCL \cite{you2020graph} are used as baselines. The experimental results of our method are obtained by incorporating our affinity uncertainty-based hardness into GraphCL to pre-train the structure2vec. The best-competing method ProGCL is adopted in the same way. The results are reported in Table~\ref{adr}. 

From the table, we can observe that: i) all three GCL methods GraphCL, ProGCL, and AUGCL can largely improve the robustness against all three graph adversarial attacks, particularly the more advanced attacks GradArgmax and RL-S2V, on different network layers, compared with the original model, ii) the robustness can be further improved by exploiting hard negative mining techniques used in ProGCL and AUGCL, compared to GraphCL, and iii) compared with ProGCL, the better hard negative mining in our method AUGCL generally results in more remarkably and stably improved robustness over the GraphCL. Overall, the proposed method AUGCL increases the classification accuracy by up to 8\% over GraphCL and up to 2.7\% over ProGCL and performs very competitively to the two methods (i.e., around 0.2\%-0.8\% difference) in the limited cases where AUGCL is not the best performer.

\subsection{Ablation Studies}
\label{ablationstudy}
This subsection evaluates the impact of using different clustering and uncertainty estimation methods in $f$ and $g$, respectively. The GraphCL$_{\text{AUGCL}}$ method is used, with GraphCL as the baseline.

\textbf{Partition Methods in $f$}.
An important module of our proposed method is the instance-wise partition function $f$. $k$-means is used by default to implement $f$. Here we also examine the use of spectral clustering \cite{ng2001spectral} to perform the binary partition. The results are shown in Table~\ref{ablation_results}. We can see that AUGCL using either spectral clustering or $k$-means achieves similar improvement over GraphCL, suggesting the stability of our method w.r.t. the generation of the partition labels. AUGCL with $k$-means clustering performs consistently better than spectral clustering. Hence, $k$-means clustering is used by default in our experiments and recommended in practice.

\textbf{Uncertainty Estimation Methods in $g$}.
The uncertainty estimation method $g$ is another important module in AUGCL. In addition to the extra class-based method used by default in AUGCL, two alternative approaches are used, including the maximum prediction probability-based method Softmax-Response \cite{geifman2017selective} and the entropy-based method Predictive Entropy \cite{lakshminarayanan2017simple}. We also include a distance-based method as another simplified variant of AUGCL. The detailed descriptions of these uncertainty estimation methods are presented in Appendix.

The results are reported in Table~\ref{ablation_results}. It is clear that regardless of the specific uncertainty estimation method used, all variants of AUGCL can generally improve the baseline GraphCL on nearly all datasets. This provides further evidence for the effectiveness of our approach. Additionally, the uncertainty estimation method matters: the default method (a recently proposed extra class-based method \cite{liu2019deep,tian2021pixel}), a more effective uncertainty estimation model than the other three methods, shows consistently better performance than the other three variants, implying that the hardness can be better captured by more advanced uncertainty estimation methods.

\subsection{Hyperparameter Analysis}
\label{hyperp}
We examine the sensitivity of AUGCL w.r.t two key hyperparameters, i.e., the uncertainty parameter $\alpha$ in (\ref{eqn:hardcl}) and the reward parameter $o$ in $\phi$ (particularly in (\ref{eqn:dg})). Without loss of generality, one graph dataset from biochemical molecules and social networks respectively, i.e., PROTEINS and IMDB-B, are used. 

\begin{figure}[htbp]
    \centering
    \subfigure[Parameter $\alpha$]{\label{p_alpha}
    \includegraphics[width=0.23\textwidth]{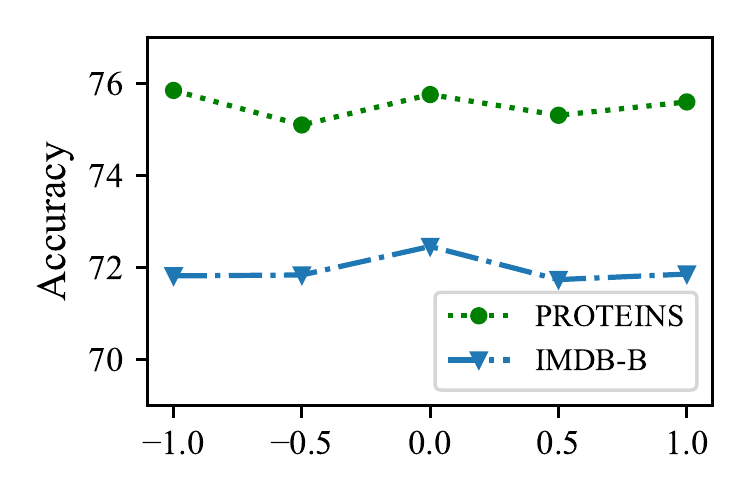}}
    \subfigure[Parameter $o$]{\label{p_o}
    \includegraphics[width=0.23\textwidth]{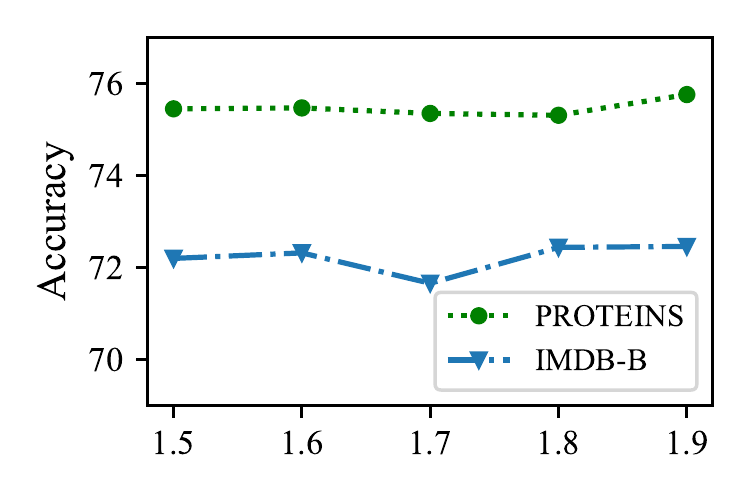}}
    \label{parameter}
    \caption{Sensitivity analysis of hyperparameters $\alpha$ and  $o$.
    } 
\end{figure}

\textbf{Uncertainty Parameter $\alpha$}.
$\alpha$ is adaptively set, depending on the uncertainty matrix $\mathbf{U}$, to enable stable performance of AUGCL. Particularly, given $\mathbf{U}$, we can calculate the mean $\mu$ and standard deviation $\delta$ of $\mathbf{U}$, based on which $\alpha$ is set to $\alpha = \frac{1}{\mu}$. We vary the parameter $\alpha$ in the range of $\{\frac{1}{\mu - \delta}, \frac{1}{\mu - 0.5\delta}, \frac{1}{\mu}, \frac{1}{\mu + 0.5\delta}, \frac{1}{\mu + \delta}\}$. The mean classification accuracy (\%) under different $\alpha$ are shown in Fig. \ref{p_alpha} where the labels in the x-axis denote the coefficient of $\delta$ when calculating $\alpha$. It is clear that the performance of our model is generally stable with varying $\alpha$, and $\alpha = \frac{1}{\mu}$ is a recommended setting.

\textbf{Reward Parameter $o$}.
We further examine the reward parameter $o$ in the uncertainty estimation model \cite{liu2019deep}. With $o$ varying in $\{1.5, 1.6, 1.7, 1.8, 1.9\}$, we report the mean classification accuracy (\%) in Fig. \ref{p_o}. The results also show that AUGCL can achieve reasonably stable performance for a wide range of the $o$ settings. 

\subsection{Training Time}
This subsection demonstrates the computation time of the proposed method. Following the experimental setting in the hyperparameter analysis section, we compare the training time on PROTEINS and IMDB-B datasets in a single run with the same setting. The comparison results of our method AUGCL to the baseline GraphCL \cite{you2020graph} and all the hardness-aware methods are reported in Table \ref{time}.

\begin{table}[h]
\caption{{Training time (second) on PROTEINS and IMDB-B.}}
\label{time}
\centering
\resizebox{0.50\textwidth}{!}{
\begin{tabular}{l|c|cccc}
\hline
Dataset & GraphCL & DCL & HCL & ProGCL & AUGRL \\
\hline
PROTEINS & 25.30& 25.82 & 26.11 & 27.15 &  29.02\\
IMDB-B & 14.22 & 14.68 & 14.78& 16.44 &18.60\\
\hline
\end{tabular}    
}
\end{table}

From the table, we can see that the utilization of hardness-aware methods leads to some increases in computation time. Although AUGCL has the highest computation time, the overall computational overhead compared to the baseline is very minor.

\subsection{Convergence}
\begin{figure}[htbp]
    \centering
    \subfigure[PROTEINS]{\label{loss_proteins}
    \includegraphics[width=0.23\textwidth]{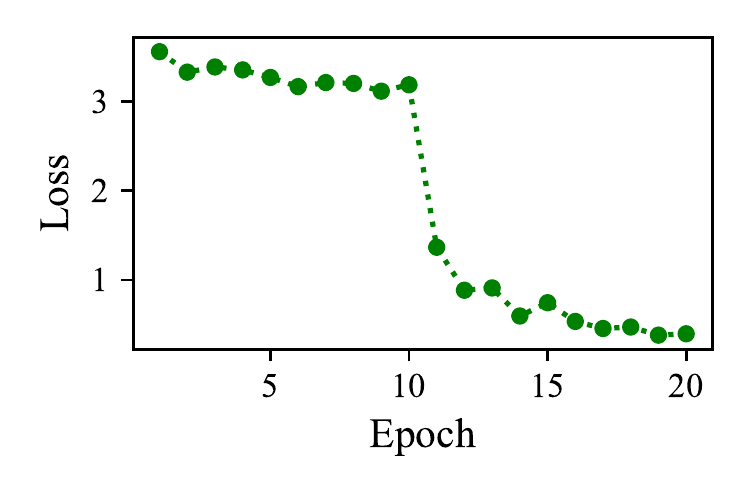}}
    \subfigure[IMDB-B]{\label{loss_imdb}
    \includegraphics[width=0.23\textwidth]{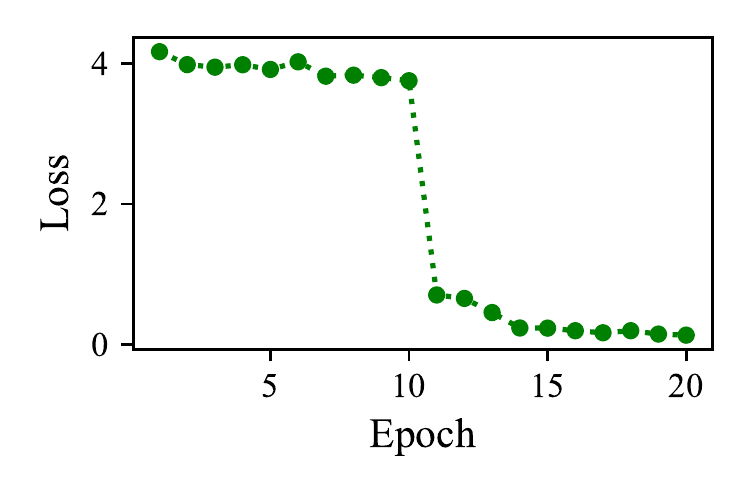}}
    \label{conver}
    \caption{Loss curve of the proposed method.} 
\end{figure}

This subsection demonstrates the convergence of the proposed method. We illustrate the loss curves on PROTEINS and IMDB-B datasets in Fig \ref{conver}.

From the figure, we can see that the proposed method achieves fast convergence. Note that there are abrupt drops for both datasets at epoch 10 instead of gradual decreases of loss values. This is attributed to the affinity uncertainty learning performed at epoch 10, in which the learned hardness-based weights effectively capture the hardness of the negatives, i.e., large weights for hard negatives while small weights for the other instances (false and easy negatives). By incorporating the learned weights, the loss from false and easy negatives is significantly reduced, resulting in the learning focusing on the hard negatives and facilitating the convergence of graph contrastive learning.

\section{Conclusion}\label{conclusion}
This paper proposes the idea of affinity uncertainty and utilizes it to measure the hardness of negative samples to improve popular GCL models. To this end, we introduce the affinity uncertainty-based hardness learning approach AUGCL that synthesizes binary partition and uncertainty estimation to learn anchor-instance-dependent hardness for all negative instances, i.e., their hardness results are relative to each anchor instance. AUGCL is a data-driven approach that eliminates the prior assumption made in very recent hardness-aware GCL methods like ProGCL \cite{xia2022progcl}, resulting in better applicability and flexibility on different graph mining tasks, as well as better robustness to diverse graph adversarial attacks. It also shows better performance in enabling different GCL loss functions, compared to a wide range of other state-of-the-art graph representation methods on graph and node classification tasks. We also show theoretically that the resulting contrastive loss in AUGCL is equivalent to a triplet loss with an adaptive margin that adaptively exploits the hard negatives with a large margin, with a small margin assigned to the other negative instances. 
 
\bibliographystyle{IEEEtran}
\bibliography{reference}

\begin{IEEEbiography}[{\includegraphics[width=1in,height=1.25in,clip,keepaspectratio]{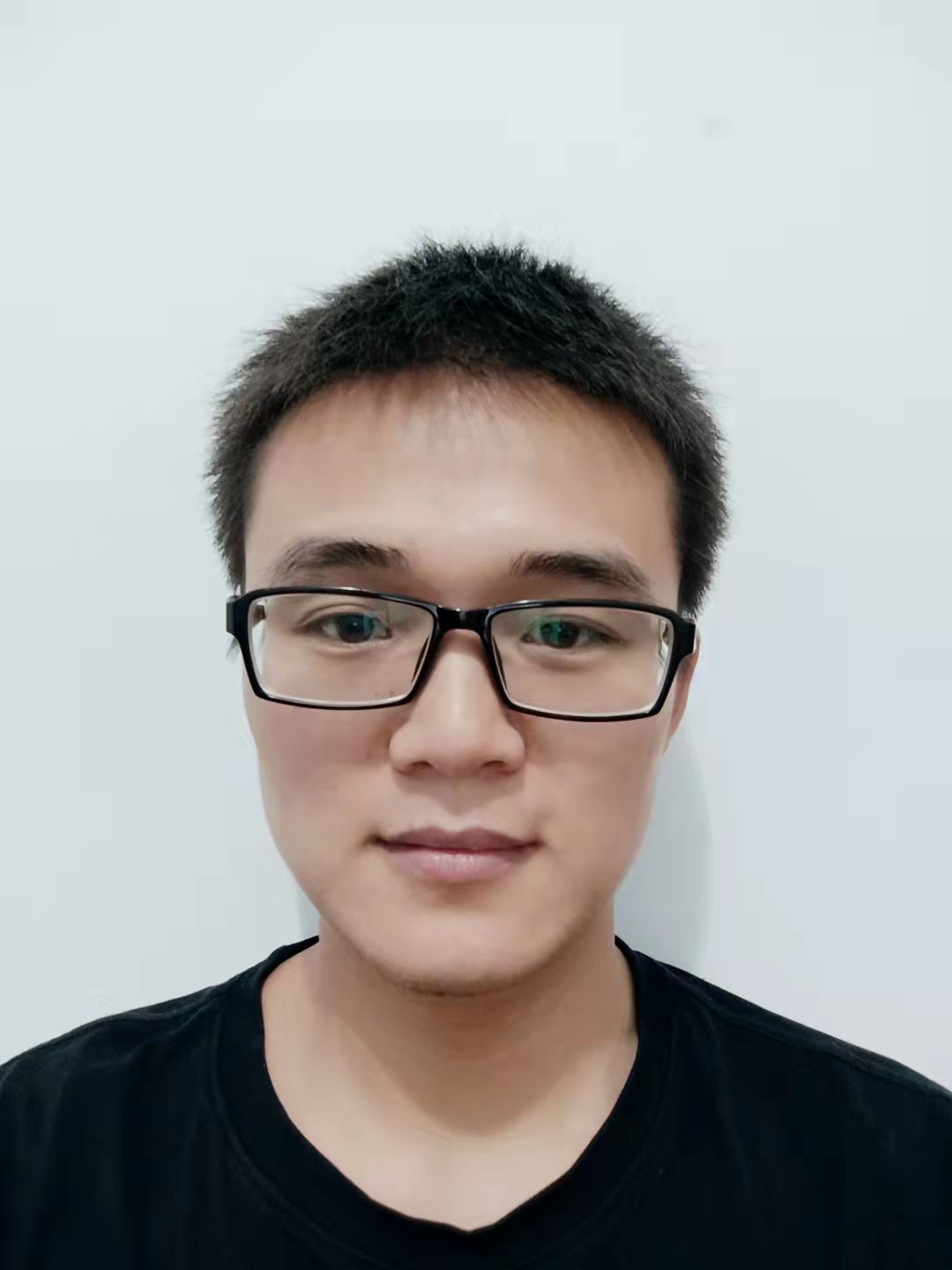}}]
{Chaoxi Niu} received the B.S. and M.S. degrees in electronic information science and technology from Lanzhou University, Lanzhou, China. He is currently pursuing the Ph.D. degree with the
University of Technology Sydney, Sydney, NSW, Australia. His current research interests primarily focus on deep learning on graphs.
\end{IEEEbiography}
\begin{IEEEbiography}[{\includegraphics[width=1in,height=1.25in,clip,keepaspectratio]{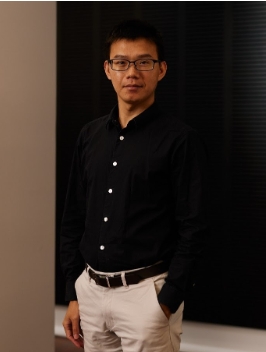}}]
{Guansong Pang} is a tenure-track Assistant Professor of Computer Science in the School of Computing and Information Systems at Singapore Management University (SMU), Singapore. Before joining SMU, he was a Research Fellow with the Australian Institute for Machine Learning (AIML). He received a Ph.D. degree from University of Technology Sydney, Australia. His research interests lie in machine
learning techniques and their applications, with a focus on handling abnormal and unknown data.
\end{IEEEbiography}
\begin{IEEEbiography}[{\includegraphics[width=1in,height=1.25in,clip,keepaspectratio]{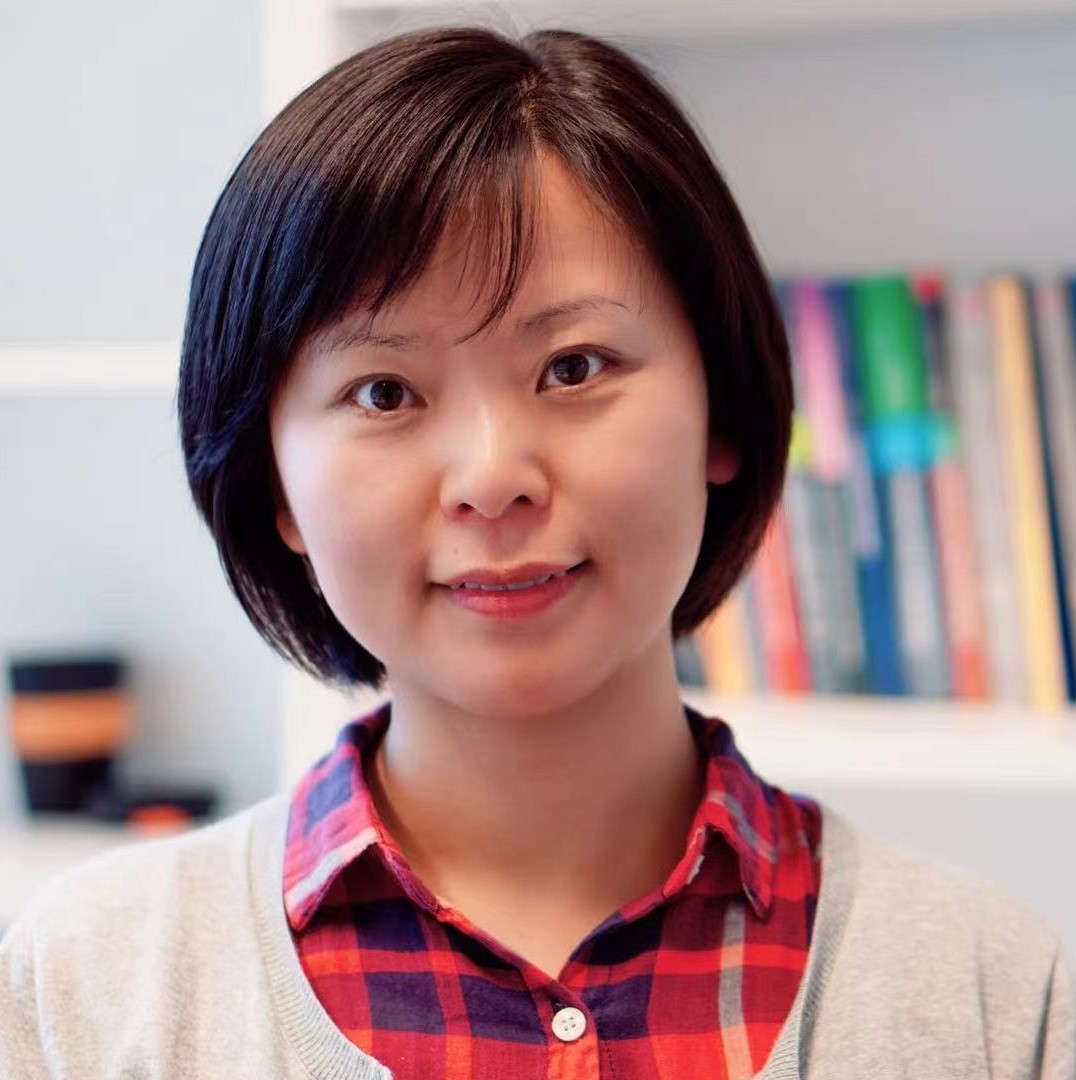}}]{Ling Chen} (Senior Member, IEEE) received the Ph.D. degree from Nanyang Technological University, Singapore, and undertook postdoctoral training at Leibniz University Hannover (L3S Research Centre), Germany. She is a Professor in the School of Computer Science at University of Technology of Sydney, Sydney, Australia. She leads the Data Science and Knowledge Discovery Laboratory (The DSKD Lab) within the Australian Artificial Intelligence Institute (AAII) at UTS. Her papers appear in major conferences and journals, including SIGKDD, AAAI, ICLR, ICDM, NeurIPS, and TNNLS. Her research interests mainly include discovering regularities and irregularities from various types of data, data representation learning, social media and social network mining, and dialogue and interactive systems.
\end{IEEEbiography}

\vfill

\end{document}